\documentclass[conference]{IEEEtran}
\usepackage{cite}
\let\citep\cite
\let\citet\cite
\usepackage{amsmath,amssymb,amsfonts}
\usepackage{algorithmic}
\usepackage{graphicx}
\usepackage{adjustbox}
\usepackage{textcomp}
\usepackage{xcolor}
\usepackage{braket}
\def\BibTeX{{\rm B\kern-.05em{\sc i\kern-.025em b}\kern-.08em
    T\kern-.1667em\lower.7ex\hbox{E}\kern-.125emX}}

\usepackage{graphicx} % Required for inserting images
\usepackage{xcolor}

\usepackage{booktabs}
\usepackage{multirow}
\usepackage{subcaption}

\usepackage{amsmath,amsthm,stmaryrd,amsfonts,amssymb}
\usepackage{braket}
\usepackage{tikzit}
\tikzstyle{dot}=[fill=black, draw=black, shape=circle, tikzit fill=black, tikzit draw=black, tikzit shape=circle, minimum size=0.02cm]

\newcommand{\define}[1]{{\bf #1}}

\theoremstyle{definition}\newtheorem{example}{Example}

\begin{document}

\title{Compositional Concept Generalization with  Variational Quantum Circuits\\
\begin{center}
\footnotesize
\textcopyright~2025 IEEE. Personal use of this material is permitted.  Permission from IEEE must be obtained for all other uses, in any current or future media, including reprinting/republishing this material for advertising or promotional purposes, creating new collective works, for resale or redistribution to servers or lists, or reuse of any copyrighted component of this work in other works.
Accepted to: \textit{2025 IEEE International Conference on Quantum Artificial Intelligence (QAI)},
Naples, Italy, 2--5 Nov 2025. DOI: to appear.
\end{center}
% \thanks{Royal Academy of Engineering}
}

%\author{\IEEEauthorblockN{Anonymous Authors}}

\author{\IEEEauthorblockN{\textsuperscript{} Hala Hawashin}
\IEEEauthorblockA{\textit{Computer Science} \\
\textit{University College London}\\
London, UK \\
hala.hawashin.23@ucl.ac.uk}
\and
\IEEEauthorblockN{\textsuperscript{} Mina Abbaszadeh}
\IEEEauthorblockA{\textit{Computer Science} \\
\textit{University College London}\\
London, UK \\
m.abbaszadeh@ucl.ac.uk}
\and
\IEEEauthorblockN{\textsuperscript{} Nicholas Joseph}
\IEEEauthorblockA{\textit{Computer Science} \\
\textit{University College London}\\
London, UK \\
nicholas.joseph.22@ucl.ac.uk}
\and
\IEEEauthorblockN{\textsuperscript{} Beth Pearson}
\IEEEauthorblockA{\textit{School of Eng. Maths. \& Tech} \\
\textit{University of Bristol}\\
Bristol, UK \\
beth.pearson@bristol.ac.uk}
\and
\IEEEauthorblockN{\textsuperscript{} Martha Lewis}
\IEEEauthorblockA{\textit{Inst. Logic Language \& Computation} \\
\textit{University of Amsterdam}\\
Amsterdam, NL \\
m.a.f.lewis@uva.nl}\\
\and
\IEEEauthorblockN{\textsuperscript{} Mehrnoosh Sadrzadeh}
\IEEEauthorblockA{\textit{Computer Science} \\
\textit{University College London}\\
London, UK \\
m.sadrzadeh@ucl.ac.uk}
}

\maketitle

\begin{abstract}
    Compositional generalization is a key facet of human cognition, but lacking in current AI tools such as vision-language models. Previous work examined whether a compositional tensor-based sentence semantics can overcome the challenge, but led to negative results.  We conjecture that the increased training efficiency of quantum models will improve performance in these tasks. We interpret the representations of compositional tensor-based models in Hilbert spaces and train  Variational Quantum Circuits to learn these representations on an image captioning task requiring compositional generalization. We used two image encoding techniques: a multi-hot encoding (MHE) on binary image vectors and an angle/amplitude encoding on image vectors taken from the vision-language model CLIP. We achieve good proof-of-concept results using noisy MHE encodings. Performance on CLIP image vectors was more mixed, but still outperformed classical compositional models.
\end{abstract}

\section{Introduction}
As humans, we are able to make sense of new situations by applying our knowledge from previously seen situations. This is called compositional generalization: having seen blue cars and red postboxes, we are able to react appropriately when we are crossing the road and a red car is speeding around the corner, even if we have never seen that particular colour of car before. In the burgeoning field of artificial intelligence and machine learning, it is essential for systems to also have this property.

The Distributional Compositional Categorical semantic model  (DisCoCat)  introduced in \cite{Coecke2010} provides an explicitly compositional way of modelling language. It maps grammatical structure, which tells us how to compose words to form phrases and sentences, to  meanings of phrases and sentences and  encodes these in vectors and higher order tensors. The grammatical type of a word dictates the structure of the vector space it inhabits, and grammatical reductions between words are modelled as tensor contraction. 

The representation of words as vectors has been used since at least \cite{salton1975vector}. In early work, the bases of the vector space in which the words are represented is interpretable, as e.g. documents \citep{salton1975vector} or other words \citep{landauer_solution_1997}, and values of vectors at each basis are derived from the statistics of words in a given document or word co-occurrence in a corpus. 
% Later work moved to a compression of the vector space via singular value decomposition \cite{deerwester} to produce dense rather than sparse vectors. Underlying these representations is the assumption that semantic similarity is interpreted as proximity in the vector space, where proximity is usually interpreted as normalized inner product. 
For a while information theoretic  functions such as local and mutual information were used to improve on the raw word statistics. Lately, machine learning via neural networks has provided substantial advances, leading to the invention of Large Language Models. \cite{mikolov_distributed_2013} is the first in this series; it introduces  word embedding methods where the dimensions of the vector space are no longer assumed to be interpretable and the notion of semantic similarity, measured by normalized inner product, becomes the only signal used to determine the value of a vector on a given basis dimension. Semantic similarity is still, however,  determined by the co-occurrence statistics of words in a given corpus. Words are initialized with two representations: as a \emph{target word} and as a \emph{context word}. Word vectors are learnt using a \emph{contrastive learning} algorithm which increases the inner product of words that occur in the same context and which decreases the inner product of words that do not occur in the same context, approximated by drawing $K$ \emph{negative samples} randomly from the vocabulary.

Specifically, the following quantity is minimized by gradient back-propagation:
\begin{equation}\label{eq:negative_sampling}
    J(\theta) = -\log \sigma(\braket{v_t|v_c}) - \sum_{k=1}^K  \log \sigma(-\braket{v_t | v_{k}})
\end{equation} 
where $\theta$ is the set of model parameters, $v_t$ is the vector of the target word, $v_c$ is the vector of the context word, $v_1, v_2, ..., v_K$ are the vector of $K$ negative samples, and $\sigma$ is the logistic learning function. This method was extended from vectors to tensors by \cite{maillard-clark-2015} and \cite{wijnholds_representation_2020} to build DisCoCat-style representations.

\paragraph{From Classical to Quantum}
 DisCoCat was conceived as a means of modelling linguistic meaning representations derived from the statistics of text corpora.  \cite{Lewis2023} extended it to multimodal meaning and developed methods for learning meaning representations from labelled images. As it is explicitly compositional, we would expect the DisCoCat representations to exhibit compositional generalization. However, its representations do not straightforwardly generalize. 
 %
 %M added
 %
 One reason for this is that DisCoCat relies on higher order tensors and tensor contraction for modelling composition. Learning  these tensors from real data and using them in concrete computations is costly on classical computers. Tensors are however natural inhabitants of quantum systems. If modelled on quantum computers, their parameters become  easier to learn and their computations  less costly. DisCoCat was inspired by Categorical Quantum Mechanics \citep{abramskybob2004} and moreover has a growing ecosystem of theory and software \citep{Kartsaklis2021, Lorenz2021} enabling the implementation of DisCoCat word and sentence representations on quantum architectures such as Variational Quantum Circuits (VQCs). Using VQCs for learning meaning representations has allowed a much more efficient implementation in linguistic tasks such as text classification, question answering, and co-reference resolution \citep{mechcoecke2020quantumnatural,Meichanetzidis_2023,WazniLoPheatSadr2024}.   In this paper, we extend these methods to multimodal cognitive tasks and conjecture that the increased efficiency of quantum computing will  help DisCoCat tensors to train better and that this will improve the performance of our proposed methods in the compositional generalization.

In order to test this conjecture, we use our quantum implementations on a spatial visual question answering task. We use a dataset developed by \cite{Lewis2023,pearson2024} in which the system must correctly identify the spatial relationship between objects in an image. We train meaning representations for the objects and relationships \emph{left} and \emph{right} and optimize these representations to describe the image. We find that the quantum trained representations learnt by VQCs outperform classically trained DisCoCat representations and perform on a par with modern multimodal transformer-based architectures.

\paragraph{From words to sentences to images}
 In section \ref{sec:discocat}, we will outline a  theoretical framework for generating the multimodal phrase and sentence vectors that we will use in this paper. Again, basis dimensions in a sentence vector space are not assumed to have any particular interpretation, and one key driver of determining the values of a sentence vector is semantic similarity as measured by inner product. So, for example, we expect a sentence vector for an image where  `kittens drink milk' to have a higher normalized inner product with images where `black cats lap water' than with those where `three children have muddy foreheads'.

In \cite{radford_learning_2021}, the contrastive learning paradigm is extended from word-word pairs as in \eqref{eq:negative_sampling} to pairs of sentence and image vectors. The assumption underlying this method is that sentence and image vectors lie in the same semantic space, and we wish to increase the similarity of images to sentences that correctly describe those images, and decrease the similarity of image vectors to sentences that do not describe those images. For example, with one negative sample, we might have 
\begin{equation}\label{eq:contrastive_images}
    J(\theta) = -\log \sigma(\braket{{\textcolor{red}{\blacksquare}}|red\_sq}) - \log\sigma(-\braket{{\textcolor{red}{\blacksquare}}|blue\_circ})
\end{equation}

In \cite{Lewis2023}, the contrastive learning paradigm is leveraged to learn compositional word representations from the images using the DisCoCat framework, which we describe in section \ref{sec:classicalexp}.

\section{Methods}

\subsection{Categorical Definitions}
\label{sec:discocat}
We review here some definitions used in the Distributional Compositional Categorial (DisCoCat) semantics. The basic idea behind DisCoCat is to model  the grammar and the meaning  of a language  in two different compact closed categories, and then set up a map from the grammar category to the meaning category in such a way that its preserves the compact closed structure. This enables the composition given by the grammatical reductions  in the grammar category to be interpreted as morphisms in the meaning category.

% \begin{dfn}
% A \define{monoidal category} is a tuple $(C, \otimes, I, \alpha, \lambda, \rho)$ where
% \begin{itemize}
% \item $C$ is a category
% \item $\otimes$, the \define{tensor}, is a functor $C\times C\rightarrow C$ where we write $A\otimes B$ for $\otimes (A,B)$
% \item $I$, the \define{unit}, is an object of $C$
% \end{itemize}
% The remaining data are natural isomorphisms, with components of type:
% \begin{itemize}
% \item $\alpha_{A,B,C} : ((A\otimes B) \otimes C) \rightarrow (A\otimes( B\otimes C))$
% \item $\rho_A : A\otimes I \rightarrow A$
% \item $\lambda_A : I\otimes A \rightarrow A$
% \end{itemize}
% These natural isomorphisms, moreover, must be such that any formal and well-typed diagram made up from $\otimes, \alpha, \lambda, \rho, \alpha^{-1}, \rho^{-1}, \lambda^{-1}$ and identities commutes. Here \lq formal\rq \:  means \lq not dependent on the structure of any particular monoidal category.\rq
% \end{dfn}

A compact closed category is a monoidal category $(C,\otimes, I)$ such that for each object $A\in C$ there are objects $A^l,A^r\in C$ (the \define{left} and \define{right duals} of $A$) and morphisms
\begin{align*}
\eta_A^l : I\rightarrow A\otimes A^l \:&\:& \eta_A^r : I\rightarrow A^r\otimes A
\\
\epsilon_A^l : A^l\otimes A \rightarrow I \:&\:& \epsilon_A^r : A\otimes A^r\rightarrow I
\end{align*}
satisfying the snake equations
\begin{align*}
&(1_A \otimes \epsilon^l )\circ (\eta^l \otimes 1_A) = 1_A &\:& (\epsilon^r \otimes 1_A) \circ (1_A \otimes \eta^r ) = 1_A
\\
&(\epsilon \otimes 1_{A^l})\circ (1_{A^l} \otimes \eta^l)= 1_{A^l} &\:& (1_{A^r} \otimes \epsilon^r ) \circ (\eta^r \otimes 1_{A^r} ) = 1_{A^r}
\end{align*}
The $\epsilon$ and $\eta$  maps are called \define{caps} and \define{cups} respectively.

As we will be modelling words, phrases, and images as (1) finite-dimensional vectors, that (2) are interpretable as states of  quantum systems, we use $\mathbf{FHilb}$ as our meaning category. This is the category of finite dimensional real inner product spaces and linear maps. The tensor $\otimes$ is the tensor product of vector spaces and $I$ is the one-dimensional space $\mathbb{R}$. Note that the basis for a space $V\otimes U$ is given by pairs $\{ v_i \otimes u_j \}_{i,j}$ of basis vectors $\{ v_i \}_i$ and $\{ u_j \}_j$ of $V$ and $U$ respectively. 
Every finite-dimensional Hilbert space is self-dual, and its cups and caps are given by
\begin{gather}
\label{eq:fhilb_cups}
\epsilon_V : V\otimes V \rightarrow \mathbb{R}::\sum_{i,j} c_{i,j} \: (v_i\otimes v_j )\mapsto \sum_{i,j} c_{i,j} \langle v_i | v_j \rangle
\\
\label{eq:fhilb_caps}
\eta_V : \mathbb{R} \rightarrow V\otimes V::1\mapsto \sum_{i} (v_i\otimes v_i )
\end{gather}

To model grammar, we use Lambek's pregroup grammars, which can be viewed as  compact closed categories.
  A \define{pregroup} is a tuple $(A,\cdot , 1, \--^l,\--^r,\leq )$ where $(A,\cdot, 1, \leq)$ is a partially ordered monoid and $\--^r, \--^l$ are functions $A\rightarrow A$ such that $\forall x\in A$, 
  \begin{align}
    \label{eq:pg}
    x\cdot x^r \leq 1 \leq x^r \cdot x\qquad x^l\cdot x\leq 1 \leq x \cdot x^l
  \end{align}
The $\cdot$ will usually be omitted, writing $xy$ for $x\cdot y$. 
We choose a set of basic linguistic types and freely generate a pregroup over these types. Words are assigned elements of the pregroup according to their grammatical type, and a string of words is assigned an element of the pregroup by applying the reductions given in \eqref{eq:pg}. The pregroup freely generated by a set $A$ is denoted by $\mathcal A$. In this paper, we use a grammar generated over the set of basic types $\mathcal B =\{n, p, s\}$, where $n$ is the type of a noun/noun phrase, $s$ is that of a sentence, and $p$ a prepositional phrase.

Pregroups can be interpreted as compact closed categories as follows. The objects of this category are the elements of the pregroup and the morphisms are given by the order structure of the pregroup. That is, there is a unique morphism~$p\rightarrow q$ if and only if~$p\leq q$. The  tensor~$\otimes$ is the monoid multiplication and the monoidal unit is the element $1$.
The left and right duals of~$p$ are~$p^l$ and~$p^r$ respectively.
The cups and caps are the unique morphisms given by the inequalities in ~\eqref{eq:pg}.

We can now define a compact closed map from the grammar category into our meaning space $\mathbf{FHilb}$ as follows. The atomic types $n$ and $s$ are mapped to Hilbert spaces $N$ and $S$. The adjoints $x^r$ and $x^l$ are both mapped to the Hilbert space dual $X^*\cong X$. The inequalities \eqref{eq:pg} corresponding to cups and caps in pregroups are mapped to the equivalent in $\mathbf{FHilb}$, as given in equations \eqref{eq:fhilb_cups} and \eqref{eq:fhilb_caps}. For full details and how this map can be seen as a strongly monoidal functor, see \cite{Coecke2010}, \cite{preller2011bell}.

Compact closed categories also have an elegant graphical calculus that allows us to depict their calculations. In this calculus, the objects of the category are interpreted as strings, and the morphisms as boxes. The composition of morphisms is given by joining strings, and the tensor product is given by placing diagrams next to each other. Finally, cups and caps are interpreted as bending wires.

\begin{figure}[!b]
    \centering
    \includegraphics[width=1\linewidth]{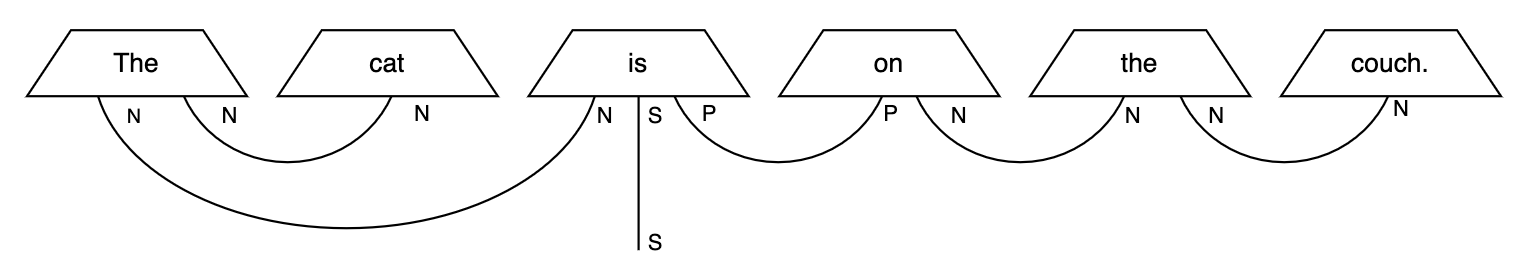}
    \caption{Sentence derivation in the graphical calculus.}
    \label{fig:t-sent}
\end{figure}

% Corrected figure above ^
% \begin{figure}[!b]
%     \centering
%    % \ctikzfig{tikzit/t-sent}
%    \includegraphics[width=0.6\textwidth]{Diagrams/predprop.png}
%     \caption{Sentence derivation in the graphical calculus\begin{color}{red} Redraw to (1) make the sames into traapezoids similar to Figure 2, and (2) change the labels  from $n$ and $n^l$ to $N$- from $s$ to $S$, from $p$ to $P$. \end{color}}
%     \label{fig:t-sent}
% \end{figure}

\begin{example}\label{ex:predprop}[Predicative Sentence with a Propositional Phrase]
In the sentence `The cat is on the couch',  the words `cat' and `couch' are given the type $n$, the preposition `on; is given the type $pn^l$,  the determiner `the' is given the types $nn^l$  and the  verb `is'  the type $n^r s p^l$. The sentence as a whole is shown to be grammatical using the inequalities below. Given suitable representations for the words `cat' and `couch' in $\mathbf{FHilb}$, this reduction is given graphically in Figure \ref{fig:t-sent}.
\begin{eqnarray*}
   (nn^l) \ n \ (n^r s p^l) \ (p n^l) \ (nn^l) \ n &\leq& \\
   n1(n^r s ) 1 1 1 &\leq& \\
   n (n^rs) 1 &\leq&\\
   1s1 &=& 1
\end{eqnarray*}

\end{example}

% \begin{example}
% For simplicity, we only use the linguistic types~$n$, for noun, and~$s$, for sentence. Hence, we will work with $\freepreg{\{n,s\}}$. Consider the sentence \textit{`Chickens cross roads.'} The nouns \textit{chickens} and \textit{roads} are of type $n$, and the transitive verb \textit{cross} is assigned the type $n^r s n^l$. \textit{`Chickens cross roads'} therefore has type~$n(n^rsn^l)n$. Then we have the following type reductions:
% \begin{align*}
%   n(n^rsn^l)n &= (nn^r)s(n^ln)\\
%   & \stackrel{(\ref{eq:pg1})}{\leq} s(n^ln) \\
%   & \stackrel{(\ref{eq:pg2})}{\leq} s 
% \end{align*}
% Note that we could also have performed these two steps in the opposite order.
% The above reduction can be given a neat graphical interpretation as follows:
% \begin{center}
% \input{tikz/chickens.tikz}  
% \end{center}
% where it is now very clear that the order of the reductions doesn't matter.
% This is a feature that is typical for pregroup grammars, while other categorial grammars such as Lambek's original categorial grammar~\cite{Lambek0} have more constraints on the order of the reductions.
% \end{example}

\subsection{Quantum Methods}
Category theory provides a structural framework to model compositional systems, which can be applied to a Hilbert space \( \mathcal{H} \)  used to formalise the computations of quantum mechanics \cite{Coecke2010}. In this framework, known as Categorical Quantum Mechanics, objects represent quantum states and are encoded as vectors \( |\psi\rangle \in \mathcal{H} \);  the inner product \( \langle \psi | \phi \rangle \) reflects the probability amplitude of measuring the state \( |\psi\rangle \) in the basis \( |\phi\rangle \) \cite{selinger2010,nielsen2010quantum}. The probability of obtaining the state \( |\phi\rangle \) when measuring \( |\psi\rangle \) is given by the square of the modulus of the inner product, \( |\langle \psi | \phi \rangle|^2 \). Operations that act on these states, such as quantum gates, are represented by morphisms, which are linear maps \( U: \mathcal{H} \to \mathcal{H} \) that describe unitary transformations between quantum states \cite{abramskybob2004}. These transformations are expressed by the below relation, where \( U \) is a unitary operator
%\begin{equation}
$|\psi'\rangle = U|\psi\rangle.$
%\end{equation}
Monoidal categories use tensor product operations to represent the interaction between multiple quantum states. The tensor product \( \otimes \) plays a crucial role in combining individual quantum states into more complex composite states. In particular, compact closed categories introduce dual objects that enable a unique reversible structure. This duality allows for left and right adjoints (denoted by \( l \) and \( r \)) to define the relationships between objects and the transformations applied to them.

\begin{figure}[h]
    \centering
   \includegraphics[width=1\linewidth]{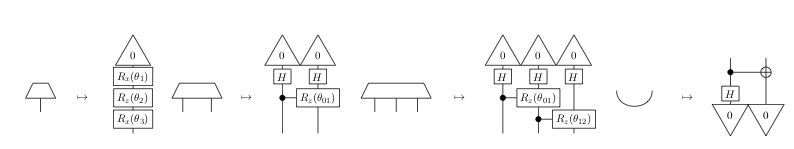}
    \caption{The IQP ansatz.}
    \label{fig:QC-Ansatz}
\end{figure}

The string diagrams of compact closed categories give a high level view of the computations we would like to perform on a quantum computer. They can be converted into Variational Quantum Circuits (VQC) by making a set of assumptions about the states of the quantum systems and their operations. These assumptions are called ans\"atze. A commonly used ansatz is the Instantaneous Quantum Polynomial (IQP). It uses Euler's decomposition to parametrize single-qubit rotations with three gates. A rank $n$ tensor is modelled by  $n$ Hadamard gates and $n-1$ entangling operations, implemented by controlled rotation gates. Cups model contractions by cancelling a type with its adjoint in the sentence structure. As a result, the corresponding qubits do not directly contribute to the final output and are post-selected to enforce the grammatical reduction. The IQP circuits for a qubit and  rank 2  and 3 tensors and a cup  are provided in Figure \ref{fig:QC-Ansatz}. 

%Two commonly used ans\"atze are the Instantaneous Quantum Polynomial (IQP) and the Sim family. Both ans\"atze use Euler's decomposition to parametrize single-qubit rotations with three gates, but there are key differences in how they handle a few multi-qubit operations, In IQP, a rank $n$ tensor has  $n$ Hadamards and $n-1$ entangling operations, making it more efficient for a fixed class of problems. In contrast, the Sim-Family ansatz  uses  $n$ single qubit rotations for a rank $n$ tensor, as well as  $kn$ entangling gates, where $k$ varies depending which element of the family is chosen. This provides greater interaction and more expressivity with better performance in a range of applications.  The IQP circuits for a rank 1, 2  and 3 tensor are provided in Figure \ref{fig:QC-Ansatz}. 

The IQP variational quantum circuit of the string diagram of  Example \ref{ex:predprop} is given below in Figure \ref{fig:VQC_Ex1}:

\begin{figure}
    \centering
    \includegraphics[width=1.0\linewidth]{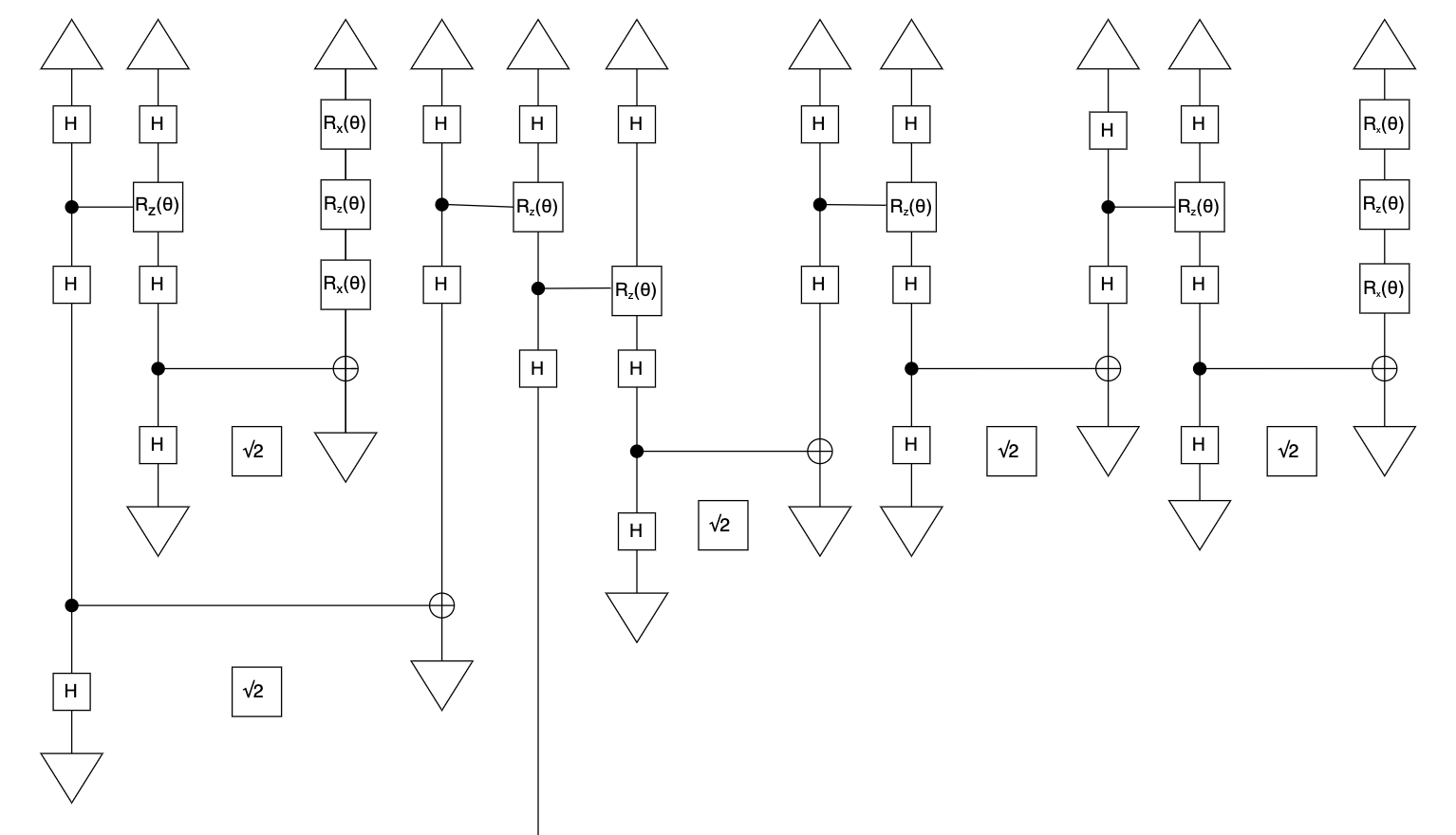}
    \caption{ A variational quantum circuit for the sentence in Example 1, which has the parse in Figure \ref{fig:t-sent}}
    \label{fig:VQC_Ex1}
\end{figure}

% Image above^
% \begin{center}
%     \begin{color}{red}
% Draw a variational quantum circuit for the sentence in Example 1, which has the parse in Figure 1,
%     \end{color}
% \end{center}

% \subsection{Classical Methods}
% may not need this. 
% \newpage
\section{Experimental setup}
We evaluate the predictions of our model on an image captioning task, where given a set of (image, caption) pairs the aim is to learn the positional relationships between the  shapes within each image and match it with the caption that correctly describes it. We use the dataset of \cite{Lewis2023}.  Our pipeline is given below:

\begin{enumerate}
\item  Compute the grammatical structure of each caption using a pregroup grammar. 
\item Compute the meaning of each caption using the  diagrammatic calculus of DisCoCat.
\item Quantum Models: \begin{itemize}
    \item Interpret the meaning diagrams as variational quantum circuits by applying the IQP ansatz to the meaning diagrams. 
    \item  Load the images onto variational quantum circuits, using  different encoding techniques, detailed in section \ref{subsec:QExp} below. 
\item In the last step, the image and sentence circuits are matched against each other and a matching score between them is computed. This matching score is used to train the parameters of the circuits on the training subset of the dataset. We test the results on the validation and test subset and provide an analysis. 
\end{itemize}
\item Classical Models: Interpret the meaning diagrams using classical DisCoCat methods from \cite{Lewis2023}, and select the sentence vector that maximizes the inner product with the image vector.
\end{enumerate} 

The quantum interpretations   are  implemented using the open source software \texttt{lambeq}. The details of the dataset that we use are given below.  The classical interpretations are implemented using methods from \cite{Lewis2023} 

%, which produces a pregroup parse \citep{Kartsaklis2021}. The software also allows one to choose  suitable representations for the words  and  computes the meanings of sentences in $\mathbf{FHilb}$ and outputs a diagram for each meaning. 

%The meaning of our above example is provided in Figure \ref{fig:String-Diagram}. 

\subsection{Dataset}
\begin{figure}[htbp]
    \centering
\includegraphics[width=0.3\linewidth]{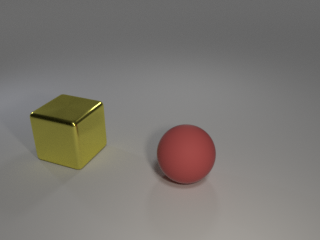}
    \caption{Correct caption: cube left sphere; Incorrect caption: cube right  sphere.}
    \label{fig:dataset_example}
\end{figure}

We use the relational split of the dataset introduced in \cite{pearson2024}. This dataset consists of images with two geometric shapes in them. The shapes are from the set $\{cube, sphere, cylinder, cone\}$. Each image is annotated with one correct caption describing the spatial relations between the shapes in the format \textit{subject relation object}, and one incorrect caption used as negative sample during training.  The incorrect captions are obtained by swapping the relation (e.g. from right to left). See Figure \ref{fig:dataset_example} for an example.  Given an image, a correct caption and an incorrect caption, the task is to decide which of the captions best describes the image. 

\begin{figure}
    \centering
    \includegraphics[width=\linewidth]{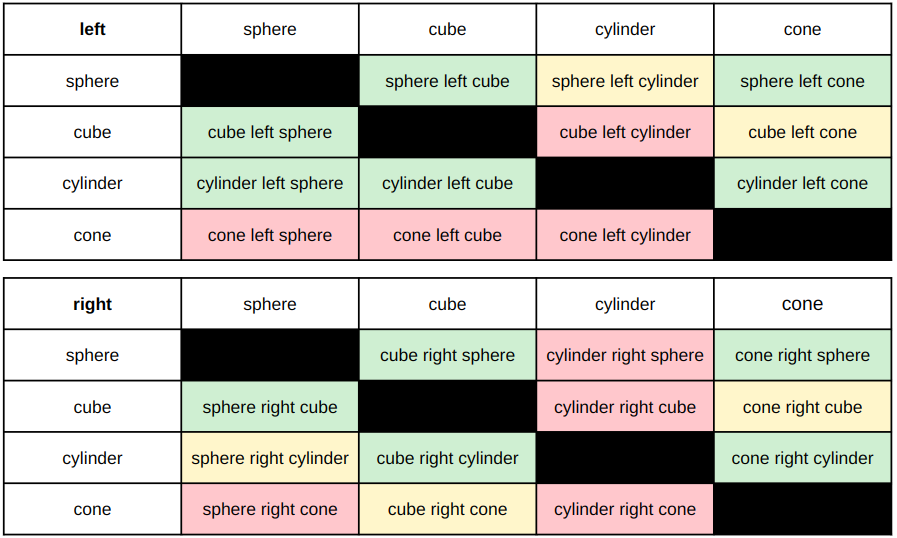}
    \caption{Dataset design. Class labels belonging to each dataset split: train and in-distribution are highlighted in green, OOD validation in yellow, and OOD test in red. Figure reproduced with permission from \cite{pearson2024}.}
    \label{fig:rel_dataset}
\end{figure}

The dataset is split into a training set, an in-distribution validation set, an out of distribution (OOD) validation set, and an OOD test set. The splits are generated as follows. The total possible captions that can be generated from $\{\textit{cube}, \textit{sphere}, \textit{cylinder}, \textit{cone}\}$ and $\{\textit{left}, \textit{right}\}$ is $4 \times 3 \times 2 = 24$. The training, OOD validation and OOD test set are given in Figure \ref{fig:rel_dataset}.

Each caption has 20 images associated with it. The model needs to learn representations of the individual words in the captions, such that when presented with an image from the OOD test set, it can compositionally generalize to the correct caption of the image.

%\begin{color}{blue} From the reviewer: It would be interesting to have a breakdown of which specific triples fell into which dataset (eg as an appendix). Was there any attention paid to whether both preposition labels/object orderings for a given object pair were in the OOD set?
%\end{color}

\subsection{Quantum experiments}
\label{subsec:QExp}

The captions of the dataset are short hand labels for the full sentences describing them. The long sentence corresponding to a caption of the form \textit{subject relation object} is `The noun is to the left/right of the noun'. As an  example consider the caption \textit{sphere left cube}, which corresponds to the sentence  `The sphere is to the left of the cube'. All these sentences have the same grammatical structure,  computed below using a pregroup grammar: 

\vspace{-.5cm}
\begin{eqnarray*}
   (nn^l) \ n \ (n^r s p^l) \ (p n^l) \ (nn^l) \ (np^l) \ (p n^l) \ (nn^l) \ n &\leq& \\
   n1(n^r s ) 1 1 1 1 1 1 &\leq& \\
   n (n^rs) 1 &\leq&\\
   1s1 &=& 1
\end{eqnarray*}

\begin{figure}[h]
    \centering
    \includegraphics[width=1\linewidth]{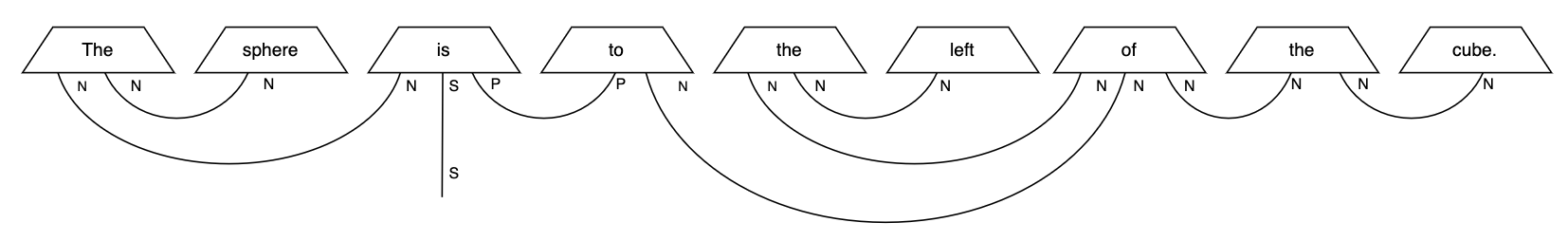}
    \caption{The diagrammatic meaning of the sentence  `The sphere is to the left of the cube'.}
    \label{fig:String-Diagram}
\end{figure}

The words within these sentences have different meaning representations and the meaning of each sentence is different from the others. The diagram corresponding to the meaning of `The cube is to the left of the sphere' is given in Figure \ref{fig:String-Diagram}.

\begin{figure}[h]
    \centering
    \includegraphics[width=0.6\linewidth]{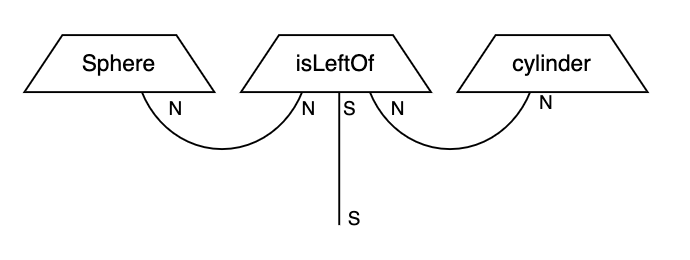}
    \caption{The  diagrammatic representation  of `Sphere isLeftOf cylinder'.}
    \label{fig:Sphere-Left-Cylinder}
\end{figure}

\begin{figure}[h]
    \centering
    \includegraphics[width=0.4\linewidth]{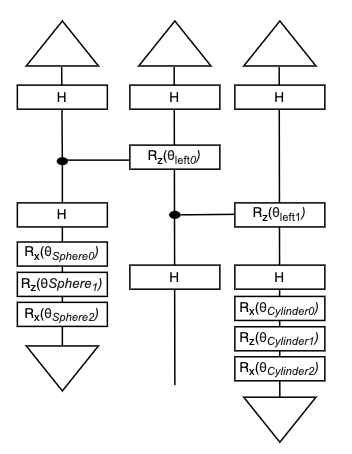}
    \caption{The VQC  of `Sphere isLeftOf cylinder'. }
    \label{fig:vqcsimple}
\end{figure}

The variational quantum circuit  of the above diagram will be complex, consisting of many rank 2 and 3 tensors  each involving  a few rotation and controlled rotation gates. A quick calculations shows this results in 27k gates per caption, where $k$  is a hyper parameter and stands for  the number of qubits used for the basic types. This results in thousands of gates  for the entire dataset. In order to decrease the circuit complexity, we simplify the sentences to the form `noun is$\{$left/right$\}$Of noun', which has all the essential semantic information of a caption and a simple grammatical reduction, as  shown below:
\[
n \ (n^r s n^l) \ n \leq 1s1 = s
\]
For example, the sentence `The cube is to the left of the sphere' is simplified to `cube isLeftOf sphere'. This forces the parser to produce simpler meaning diagrams that result in less complex circuits. For instance, the meaning of `sphere isLeftOf cylinder' becomes the diagram  of Figure \ref{fig:Sphere-Left-Cylinder}, which is translated into the circuit of Figure \ref{fig:vqcsimple}.

Two different encodings are used for the image vectors:  
\begin{enumerate}
    \item {\bf Multi-Hot Encodings (MHE).} This method turns the information of an image into a binary vector. Given two shapes: a Shape$_1$ and a Shape$_2$, and a relation between them, if the relation is `isLeftOf', the sentence vector is   of the form [Shape$_1$, Shape$_2$] and if it is `isRightOf', the sentence vectors  is  of the form [Shape$_2$, Shape$_1$]. Shape$_1$ and Shape$_2$ are encoded in a binary notation,  where each one is represented by  4 bits , resulting in an 8  bits vectors for the whole caption. The resulting embeddings are shown in tables \ref{tab:imageOHE-shape} and \ref{tab:imageOHE-preposition}. MHE's focus on the essential data required to solve a task and discard other details typically captured by classical image encoders, such as colour and variations in size. They thus serve as a proof-of-concept for our quantum model.
     \item {\bf CLIP Encodings.} Here we use the image vectors learnt by  OpenAI's Transformer-based vision language model CLIP  \citep{radford_learning_2021}.  These vectors have 512 dimensions and   include all the data of an image, including sizes and colours of the shapes, into account. They are harder to reason about. 
    We reduce the dimensions of the CLIP image vectors  by using Principal Component Analysis (PCA), yielding a low-dimensional feature vector. The size of these feature vectors is a hyper-parameter of the model; we used 3, 9 and 12 qubits.   The feature  vectors are  loaded into quantum circuits using  angle and amplitude   techniques  developed in \cite{Kerenidis_2020}.  
     
     \underline{Amplitude encoding} embeds the normalized $2^n$-dimensional feature vector into the amplitudes of a quantum state over $n$ qubits. This creates a superposition $\sum_i x_i \ket{i}$, where classical data is globally encoded and must be processed by a quantum model to yield useful outcomes. 
In \underline{angle encoding}, the  features are  mapped to the rotation angles of controlled quantum gates, applied across pairs of qubits. The control-target structure enables conditional rotations, introducing entanglement and enhancing the circuit’s expressive capacity. 

\end{enumerate}

\begin{table}[h]
\centering
   \hspace{-0.7cm} \begin{minipage}{3cm}
        \centering
        \begin{tabular}{lc}
            \hline
            \multicolumn{2}{c}{\textbf{Shapes}} \\
            \hline
            Cylinder & [1, 0, 0, 0] \\
            Sphere   & [0, 1, 0, 0] \\
            Cube     & [0, 0, 1, 0] \\
            Cone     & [0, 0, 0, 1] \\
            \hline
        \end{tabular}
        \caption{One-hot embeddings (OHE) for shapes.}
        \label{tab:imageOHE-shape}
    \end{minipage} \qquad  
    \begin{minipage}{4cm}
        \centering
        \begin{tabular}{lc}
            \hline
            \multicolumn{2}{c}{\textbf{Sentences}} \\
            \hline
             X Left-of Y   & [$\underbrace{a, b, c, d}_{\mbox{OHE of X}}$, $\underbrace{a', b', c', d'}_{\mbox{OHE of Y}}$] \\
            X Right-of Y &[$\underbrace{a', b', c', d'}_{\mbox{OHE of Y}}$, $\underbrace{a, b, c, d}_{\mbox{OHE of X}}$] \\
            \hline
        \end{tabular}
        \caption{MHE vectors for sentences.}
        \label{tab:imageOHE-preposition}
    \end{minipage}
\end{table}

\begin{figure}[h]
    \centering
    \includegraphics[width=0.7\linewidth]{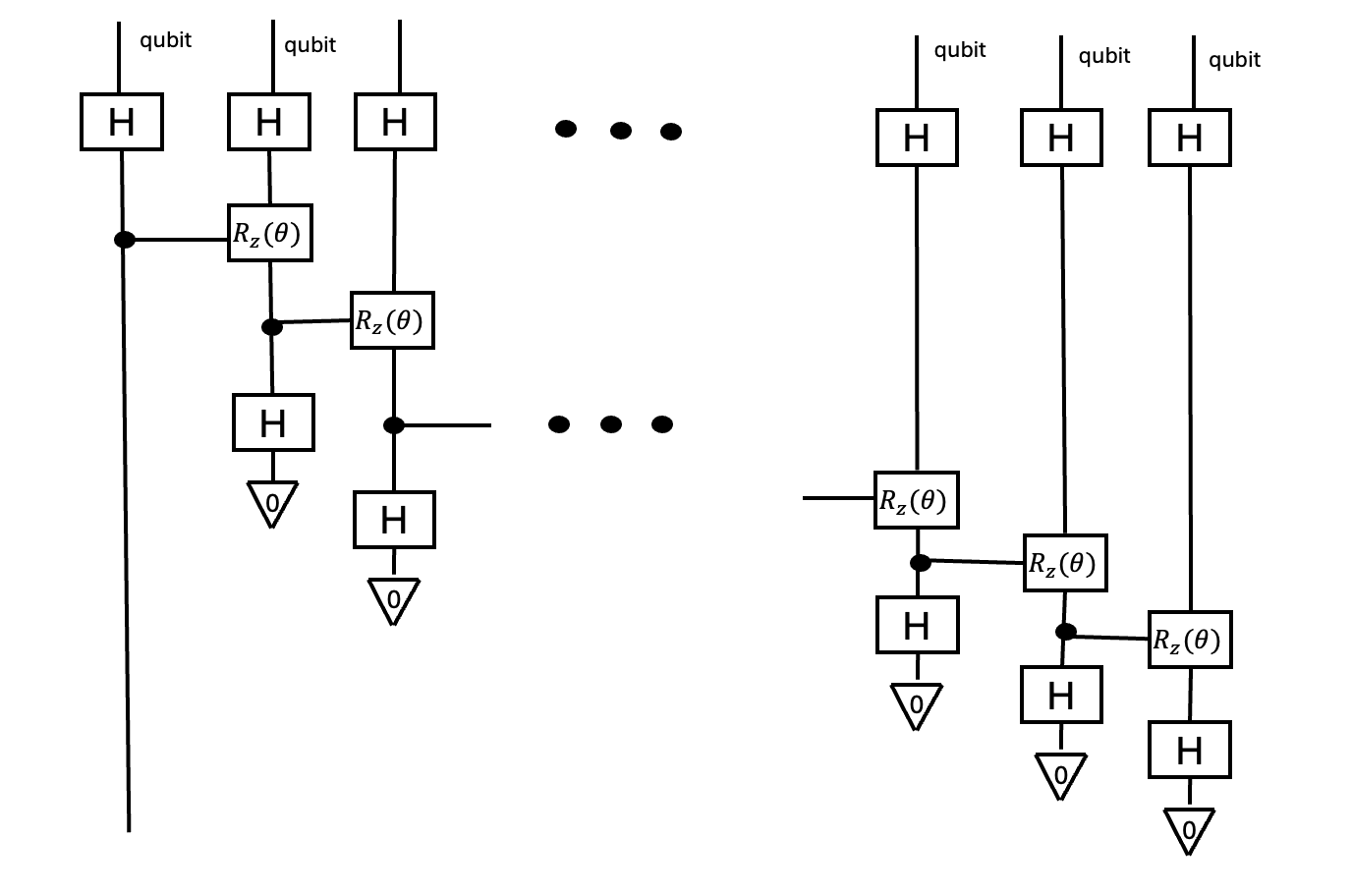}
    \caption{The trainable box.}
    \label{fig:box}
\end{figure}

A matching score is computed between each  (image, caption) pair by taking the output of their quantum circuits and evaluating the inner product between them. To compute the inner product between the image output and the sentence output, they have to have the same dimensionality. To achieve this, we implemented two strategies: 
\begin{enumerate}
    \item  {\bf Unifying Trainable Box.} This is added at the end of the quantum circuit of the image and reduces the image encoding to a single qubit, see Figure \ref{fig:box}.  The box has an internal circuit that follows the structure of an IQP ansatz, consisting of controlled rotations gates for entanglement followed by measurements with post selection. The angles of the gates are trained, which means that we are in effect  training the image vectors.  
    \item {\bf Higher Dimensional Sentence Spaces.} In this method, we increase the number of qubits of the sentence  spaces to match the number of qubits of the image vectors. No rotation gates are added and  the image vectors are not trained. 
\end{enumerate}
%Otherwise, we need to increase the number of qubits in the caption circuit to match the dimensionality of the image circuit, enabling the computation of cosine similarity between their output vectors.

\subsection{Classical experiments}
\label{sec:classicalexp}
\begin{figure}[h]
    \centering
    \includegraphics[width=0.7\linewidth]{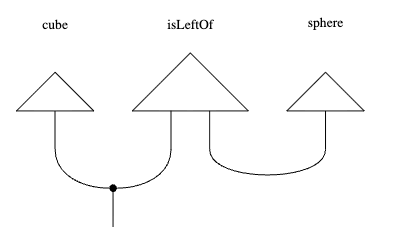}
    \caption{Diagram of the subject, relation, and object representations for the classical representations.}
    \label{fig:copy-subj}
\end{figure}

We also implement the classical DisCoCat representations of relations and sentences using techniques from \cite{Lewis2023}. 
%In these representations, 768 dimensional image embeddings are extracted from CLIP. 
In these representations the  relations \emph{isLeftOf} and \emph{isRightOf} are  represented as matrices in $\mathbb{R}^{d\times d}$, where $d$ is the dimension of the object vectors, and composed with the subject and object using the Copy-Subj methodology from \cite{kartsaklis2013reasoning}, illustrated in Figure \ref{fig:copy-subj}. 

%\tikzset{tikzfig/.append style={scale=0.8, transform shape}}

% —OR— only for this one figure:
%{%
%  \tikzset{tikzfig/.append style={scale=1.5, transform shape}}
% \begin{figure}[h]
%     \centering
%     \ctikzfig{tikzit/copy-subj}
%     \caption{Diagram of the subject, relation, and object representations for the classical representations.}
%     \label{fig:copy-subj}
% \end{figure}}

Subject, object and relation representations are implemented in PyTorch and trained using contrastive learning. Specifically, we minimize the following quantity using gradient descent

\begin{align*}
        \label{eq:rel}
   J(\theta) = 
    &-\log \sigma(\braket{m|subj\odot(\underline{rel}\cdot obj)})  
    \\&-    \log \sigma(-\braket{m|subj\odot(\underline{rel}_{opp}\cdot obj)})
\end{align*}

Here,  $m$ is the vector for the image, `subj rel obj' is the correct textual description for the image, `subj rel$_{opp}$ obj' is the incorrect textual description for the image,  and $subj\odot(\underline{rel }\cdot obj)$ forms the sentence vector for the corresponding label. We train a classical DisCoCat model for both the MHE sentence vectors described in table \ref{tab:imageOHE-preposition} and for image vectors extracted from CLIP.

\section{Results and Discussions}

The results are presented in Tables \ref{tab:rvl_classification1} and \ref{tab:rvl_classification2}. Each quantum model was trained for 100 epochs  with a learning rate of 0.001, a seed of 1 and a batch size of 8. We used 9 qubits for the Angle encoding and 12 qubits for the Amplitude encoding.

Classical models were trained for 50 epochs. Classic-CLIP was trained with a learning rate of $10^{-5}$. After hyperparameter selection, the Classic-DisCoCat model was trained with learning rate 0.1 for CLIP vectors and 0.01 for MHE vectors.

Acorss all models, results were collected for 4 random seeds and the results corresponding to the seed with the best OOD validation accuracy reported.

\begin{table}[!h]
    \centering
    \renewcommand{\arraystretch}{1.3}
    \setlength{\tabcolsep}{7pt}
    \begin{tabular}{l l c c c}
        \multicolumn{5}{c}{\textbf{Results with a Unifying  Trainable Image Box}} \\ 
        \toprule
        \textbf{Models} & \textit{Method} & \textbf{Train} & \textbf{Valid} & \textbf{Test} \\
        \midrule
        \textbf{Quantum-MHE} & with noise & 93.86\% & 67.00\% & 64.06\% \\
        \textbf{Quantum-MHE} & without noise & 100.00\% & 70.00\% & 62.50\% \\
      %  & Sim14 & 60.91\% & 46.00\% & 55.62\% \\
        \midrule
        \textbf{Quantum-CLIP} & Angle Enc.  & 79.09\% & 61.50\% & 50.31\% \\
        \textbf{Quantum-CLIP} & Amplitude Enc. & 60.45\% & 41.50\% & 41.25\% \\
  %      & Sim14 & 68.41\% & 34.00\% & 51.88\% \\
        \midrule
        \textbf{Classic-CLIP} & Frozen & 45.91\% & 48.00\% & 62.50\% \\
                              & Fine-tuned & 90.91\% & 63.00\% & 70.00\% \\
        \midrule
        \textbf{Classic-DisCoCat} & CLIP & 100.00\% & 66.00\% & 0.00\%\\
        & MHE-noise & 100.00\% & 60.00\% & 30.63\%\\
        
        \bottomrule
    \end{tabular}
    \caption{Results of experiments.}
    \label{tab:rvl_classification1}
\end{table}

\begin{table}[!h]
    \centering
    \renewcommand{\arraystretch}{1.3}
    \setlength{\tabcolsep}{7pt}
    \begin{tabular}{l l c c c}
        \multicolumn{5}{c}{\textbf{Results by Increasing the Sentence Dimensions}} \\ 
        \toprule
        \textbf{Models} & \textit{Method} & \textbf{Train} & \textbf{Valid} & \textbf{Test} \\
        \midrule
        \textbf{Quantum-MHE} & with noise & 86.82\% & 60.00\% & 50.00\% \\
        \textbf{Quantum-MHE} & without noise & 87.05\% & 60.00\% & 50.00\% \\
      %  & Sim14 & 60.91\% & 46.00\% & 55.62\% \\
        \midrule
        \textbf{Quantum-CLIP} & Angle Enc.  & 82.73\% & 62.00\% & 48.75\% \\
        \textbf{Quantum-CLIP} & Amplitude Enc. & 49.09\% & 49.50\% & 46.56\% \\
  %      & Sim14 & 68.41\% & 34.00\% & 51.88\% \\
        \midrule
        \textbf{Classic-CLIP} & Frozen & 45.91\% & 48.00\% & 62.50\% \\
                              & Fine-tuned & 90.91\% & 63.00\% & 70.00\% \\
        \midrule
        \textbf{Classic-DisCoCat} & CLIP & 100.00\% & 66.00\% & 0.00\%\\
        & MHE-noise & 100.00\% & 60.00\% & 30.63\%\\
        \bottomrule
    \end{tabular}
    \caption{Results of experiments.}
    \label{tab:rvl_classification2}
\end{table}

% %cmd+?
% \begin{table}[!h]
%     \centering
%     \renewcommand{\arraystretch}{1.3}
%     \setlength{\tabcolsep}{8pt}
%     \begin{tabular}{l l c c c}
%         \multicolumn{5}{c}{\textbf{OLD TABLE: Preposition-RL}} \\ 
%         \toprule
%         \textbf{Models} & \textit{Method} & \textbf{Train} & \textbf{Valid} & \textbf{Test} \\
%         \midrule
%         \textbf{Quantum-MHE} & IQP & 70.91\% & 52.00\% & 56.25\% \\
%         & Sim14 & 60.91\% & 46.00\% & 55.62\% \\
%         \midrule
%         \textbf{Quantum-CLIP} & IQP & 65.91\% & 53.00\% & 50.62\% \\
%         & Sim14 & 68.41\% & 34.00\% & 51.88\% \\
%         \midrule
%         \textbf{Classic-CLIP} & Frozen & 45.91\% & 48.00\% & 62.50\% \\
%                               & Fine-tuned & 90.91\% & 63.00\% & 70.00\% \\
%         \midrule
%         \textbf{Classic-DisCoCat}$^*$ & - & 99.00\% & 0\% & 0\%\\
%         \bottomrule
%     \end{tabular}
%     \caption{Results of experiments conducted on the \textit{RightvsLeft} classification task, evaluated on the training, OOD validation, and OOD test sets. Two quantum approaches for image vector embeddings were tested using Sim14 and IQP ansätze: One-Hot-Encoding (MHE) and CLIP. A classical CLIP model was evaluated with two methods: Frozen and Fine-tuned. Note that Classic-DisCoCat results are for an alternative, more complex task that encompasses both Preposition-RL and Logical Swap}
%     \label{tab:rvl_classification}
% \end{table}

We see that overall, using the Unifying Trainable Image Box (Table \ref{tab:rvl_classification1}) gives the best performance for quantum models. Unsurprisingly, quantum models with multi-hot image encodings (Quantum-MHE, 64.06\% test accuracy) give the best results across quantum models. This is expected, since these image embeddings are designed to be well-separated and have appropriate similarity relations by class. Notably, Quantum-MHE performs more strongly than Classical-DisCoCat with MHE. We see that Classical-DisCoCat with MHE achieves strong training performance, but fails to generalize effectively, with only 30.36\% test accuracy.

Over the quantum models trained with CLIP vectors, we see that the angle encoding generally performs more strongly than the amplitude encoding. On the test set, Quantum-CLIP achieves around baseline accuracy (highest test accuracy 50.31\%). However, this is substantially better than Classic-DisCoCat trained with CLIP vectors, where the model overfits to the training data and achieves 0\% accuracy on the test split. This is in line with results previously found in \cite{Lewis2023}.

The CLIP model itself already performs strongly on the test set (62.5\% test accuracy), and this improves after fine-tuning. However, the CLIP mode is pretrained on a large volume of data, and has many orders of magnitude more parameters than our quantum models (tens of millions for CLIP vs. hundreds for our quantum models). We note that on the unseen validation set, our performance (e.g. 61.5\% for Quantum-CLIP with angle encoding) approaches that of fine-tuned CLIP (63\%).

% The  Quantum-MHE achieved the best results, scoring an accuracy of 64.06\% in the noiseless model and of 62.50\% in the noisy model. These results were much higher than that of the classical DisCoCat models, which overfits to the training data, attaining 0\% on the test split. This shows that the compositional theory prescribed by DisCoCat is much more efficiently trainable using quantum methods than the classical methods and proves the purpose of this paper.   

% The frozen CLIP model demonstrated moderate performance on the test set (62.50\%), which is higher than its performance on the training split  (45.91\%) and validation split (48.00\%). This shows that the frozen CLIP embeddings do not efficiently encode obects and attributes. %This limitation likely extends to the quantum methods using CLIP embeddings, as these methods would have the same techniques in feature extraction, resulting in poor generalization. 
% When fine-tuned, the CLIP model did achieve the highest overall performance. This was expected as the  datasets are  small and fine tuning is a power procedure employing thousands of parameters.  Still, it does not help us conclude that CLIP can learn concept generalisation. As when the particularities of the task changes, new fine tuning procedures are required which goes against the nature of the learning. 

Surprisingly, the use of CLIP image vectors when training the DisCoCat circuits did not perform as well as expected. Although still better than classical DisCoCat, neither training a unifying box for images, not increasing the sentence dimensions achieved a good results. An error analysis revealed interesting results. The model struggled in learning the shapes: sphere was not recognised at all (0\% of the time). As a result, whenever sphere was to the left or right of any other shape, the whole sentence was not recognised correctly, leading to a 0\% performance whenever sphere was to the left or right of another shape.  The recognition of other shapes was better, but still not as high. We conclude that a  staged training procedure can improve this, where first we train the features of shapes and  only then the relationships between them.  
In our quantum experiments, each caption is encoded by a variational circuit with 1 qubit per NOUN and 3 layers, yielding 36 trainable parameters per caption, fewer than classical models.
% We also observe that the Quantum model demonstrates performance approaching that of the fine-tuned classical CLIP, despite using many orders of magnitude fewer parameters (hundreds vs tens of millions). Interestingly, both models also exhibited a similar pattern of results where the gap between validation and test set was small. This stability on unseen data shows that both models were able to better understand the underlying structure and generalize.

\section{Conclusions}
We  examined the performance of quantum natural language representations on an image labelling task requiring compositional generalization. We found that while the performance of the quantum methods are still lacking, they consistently outperform a classically trained model, indicating that the quantum models are less susceptible to overfitting on the training set. While the modern Transformer-based architecture CLIP performs best overall, that model has a substantial advantage in terms of pretraining data and model size. We show that on a proof-of-concept task, our quantum model can begin to generalize to out-of-distribution inputs. The choice of implementation has a large effect on the performance of quantum methods, and future work could include a systematic analysis of the types of encoding and circuit that are most useful for multimodal architectures. We also found that models were less able to learn certain shapes, meaning that an analysis of training methods is necessary to improve performance on these compositional generalization tasks.

\bibliographystyle{IEEEtran}
\bibliography{qpl_multimodality,generic}

\begin{thebibliography}{10}
\providecommand{\url}[1]{#1}
\csname url@samestyle\endcsname
\providecommand{\newblock}{\relax}
\providecommand{\bibinfo}[2]{#2}
\providecommand{\BIBentrySTDinterwordspacing}{\spaceskip=0pt\relax}
\providecommand{\BIBentryALTinterwordstretchfactor}{4}
\providecommand{\BIBentryALTinterwordspacing}{\spaceskip=\fontdimen2\font plus
\BIBentryALTinterwordstretchfactor\fontdimen3\font minus \fontdimen4\font\relax}
\providecommand{\BIBforeignlanguage}[2]{{%
\expandafter\ifx\csname l@#1\endcsname\relax
\typeout{** WARNING: IEEEtran.bst: No hyphenation pattern has been}%
\typeout{** loaded for the language `#1'. Using the pattern for}%
\typeout{** the default language instead.}%
\else
\language=\csname l@#1\endcsname
\fi
#2}}
\providecommand{\BIBdecl}{\relax}
\BIBdecl

\bibitem{Coecke2010}
\BIBentryALTinterwordspacing
B.~Coecke, M.~Sadrzadeh, and S.~Clark, ``Mathematical foundations for a compositional distributional model of meaning,'' \emph{arXiv:1003.4394 [cs, math]}, Mar. 2010. [Online]. Available: \url{http://arxiv.org/abs/1003.4394}
\BIBentrySTDinterwordspacing

\bibitem{salton1975vector}
G.~Salton, A.~Wong, and C.~S. Yang, ``A vector space model for automatic indexing,'' \emph{Communications of the ACM}, vol.~18, no.~11, pp. 613--620, 1975.

\bibitem{landauer_solution_1997}
T.~K. Landauer and S.~T. Dumais, ``A solution to {Plato}'s problem: the latent semantic analysis theory of acquisition, induction, and representation of knowledge,'' \emph{Psychological Review}, vol. 104, no.~2, pp. 211--240, 1997.

\bibitem{mikolov_distributed_2013}
T.~Mikolov, I.~Sutskever, K.~Chen, G.~S. Corrado, and J.~Dean, ``Distributed representations of words and phrases and their compositionality,'' in \emph{Advances in Neural Information Processing Systems (NeurIPS)}, vol.~26.\hskip 1em plus 0.5em minus 0.4em\relax Curran Associates, Inc., 2013, pp. 3111--3119.

\bibitem{maillard-clark-2015}
\BIBentryALTinterwordspacing
J.~Maillard and S.~Clark, ``Learning adjective meanings with a tensor-based skip-gram model,'' in \emph{Proceedings of the Nineteenth Conference on Computational Natural Language Learning}.\hskip 1em plus 0.5em minus 0.4em\relax Beijing, China: Association for Computational Linguistics, Jul. 2015, pp. 327--331. [Online]. Available: \url{https://aclanthology.org/K15-1035/}
\BIBentrySTDinterwordspacing

\bibitem{wijnholds_representation_2020}
\BIBentryALTinterwordspacing
G.~Wijnholds, M.~Sadrzadeh, and S.~Clark, ``Representation learning for type-driven composition,'' in \emph{Proceedings of the 24th Conference on Computational Natural Language Learning (CoNLL)}.\hskip 1em plus 0.5em minus 0.4em\relax Online: Association for Computational Linguistics, Nov. 2020, pp. 313--324. [Online]. Available: \url{https://aclanthology.org/2020.conll-1.24}
\BIBentrySTDinterwordspacing

\bibitem{Lewis2023}
M.~Lewis, N.~Nayak, P.~Yu, J.~Merullo, Q.~Yu, S.~Bach, and E.~Pavlick, ``Does {CLIP} bind concepts? probing compositionality in large image models,'' in \emph{Findings of the Association for Computational Linguistics: EACL 2024}, Y.~Graham and M.~Purver, Eds.\hskip 1em plus 0.5em minus 0.4em\relax St. Julian's, Malta: Association for Computational Linguistics, Mar. 2024, pp. 1487--1500.

\bibitem{abramskybob2004}
\BIBentryALTinterwordspacing
S.~Abramsky and B.~Coecke, ``A categorical semantics of quantum protocols,'' 2007. [Online]. Available: \url{https://arxiv.org/abs/quant-ph/0402130}
\BIBentrySTDinterwordspacing

\bibitem{Kartsaklis2021}
\BIBentryALTinterwordspacing
D.~Kartsaklis, I.~Fan, R.~Yeung, A.~Pearson, R.~Lorenz, A.~Toumi, G.~de~Felice, K.~Meichanetzidis, S.~Clark, and B.~Coecke, ``Lambeq: An efficient high-level python library for quantum {NLP},'' \emph{arXiv:2110.04236 [quant-ph]}, Oct. 2021. [Online]. Available: \url{http://arxiv.org/abs/2110.04236}
\BIBentrySTDinterwordspacing

\bibitem{Lorenz2021}
\BIBentryALTinterwordspacing
R.~Lorenz, A.~Pearson, K.~Meichanetzidis, D.~Kartsaklis, and B.~Coecke, ``{QNLP} in practice: running compositional models of meaning on a quantum computer,'' \emph{arXiv:2102.12846 [quant-ph]}, Feb. 2021. [Online]. Available: \url{http://arxiv.org/abs/2102.12846}
\BIBentrySTDinterwordspacing

\bibitem{mechcoecke2020quantumnatural}
\BIBentryALTinterwordspacing
B.~Coecke, G.~de~Felice, K.~Meichanetzidis, and A.~Toumi, ``Foundations for near-term quantum natural language processing,'' 2020. [Online]. Available: \url{https://arxiv.org/abs/2012.03755}
\BIBentrySTDinterwordspacing

\bibitem{Meichanetzidis_2023}
K.~Meichanetzidis, A.~Toumi, G.~de~Felice, and B.~Coecke, ``Grammar-aware sentence classification on quantum computers,'' \emph{Quantum Machine Intelligence}, vol.~5, no.~1, Feb. 2023.

\bibitem{WazniLoPheatSadr2024}
\BIBentryALTinterwordspacing
H.~Wazni, K.~I. Lo, L.~McPheat, and M.~Sadrzadeh, ``Large scale structure-aware pronoun resolution using quantum natural language processing,'' \emph{Quantum Machine Intelligence}, vol.~6, p.~60, 2024. [Online]. Available: \url{https://doi.org/10.1007/s42484-024-00193-w}
\BIBentrySTDinterwordspacing

\bibitem{pearson2024}
B.~Pearson, M.~Wray, and M.~Lewis, ``Diffusion models for improved compositional generalisation in {VLMs},'' Proc. 2nd Workshop on What is Next in Multimodal Foundation Models?, 2024.

\bibitem{radford_learning_2021}
A.~Radford, J.~W. Kim, C.~Hallacy, A.~Ramesh, G.~Goh, S.~Agarwal, G.~Sastry, A.~Askell, P.~Mishkin, J.~Clark, G.~Krueger, and I.~Sutskever, ``Learning transferable visual models from natural language supervision,'' in \emph{Proceedings of the 38th International Conference on Machine Learning (ICML)}, ser. Proceedings of Machine Learning Research, vol. 139.\hskip 1em plus 0.5em minus 0.4em\relax PMLR, Jul. 2021, pp. 8748--8763.

\bibitem{preller2011bell}
A.~Preller and M.~Sadrzadeh, ``Bell states and negative sentences in the distributed model of meaning,'' \emph{Electronic Notes in Theoretical Computer Science}, vol. 270, no.~2, pp. 141--153, 2011.

\bibitem{selinger2010}
\BIBentryALTinterwordspacing
P.~Selinger, ``A survey of graphical languages for monoidal categories,'' in \emph{New Structures for Physics}.\hskip 1em plus 0.5em minus 0.4em\relax Springer, 2010, pp. 289--355. [Online]. Available: \url{https://arxiv.org/abs/0908.3347}
\BIBentrySTDinterwordspacing

\bibitem{nielsen2010quantum}
M.~A. Nielsen and I.~L. Chuang, \emph{Quantum Computation and Quantum Information}.\hskip 1em plus 0.5em minus 0.4em\relax Cambridge, UK: Cambridge University Press, 2010.

\bibitem{Kerenidis_2020}
I.~Kerenidis and A.~Luongo, ``Classification of the {MNIST} data set with quantum slow feature analysis,'' \emph{Phys. Rev. A}, vol. 101, no.~6, p. 062327, Jun. 2020.

\bibitem{kartsaklis2013reasoning}
D.~Kartsaklis, M.~Sadrzadeh, S.~Pulman, and B.~Coecke, ``Reasoning about meaning in natural language with compact closed categories and {Frobenius} algebras,'' in \emph{Logic and Algebraic Structures in Quantum Computing}, 2013, p. 199.

\end{thebibliography}

\end{document}